\documentclass[letterpaper, 10 pt, conference]{ieeeconf}  

\IEEEoverridecommandlockouts                              

\usepackage{amsmath}
\usepackage{amsfonts}
\usepackage { graphicx } 
\usepackage { subfigure } 
\graphicspath {  { ./images/ }  }
\usepackage{caption}
\usepackage{subcaption}
\usepackage[utf8]{inputenc}
\usepackage{threeparttable}
\usepackage{hyperref}
\usepackage{cite}

\title{\LARGE \bf
Efficient Collision Detection Framework\\ for Enhancing Collision-Free Robot Motion
}

\author{Xiankun Zhu, Yucheng Xin, Shoujie Li, Houde Liu{$^\dagger$},  Chongkun Xia, Bin Liang
\thanks{This work was supported by National Natural Science Foundation of China (No. 62203260, 92248304), The Shenzhen Science Fund for Distinguished Young Scholars (RCJC20210706091946001), Tsinghua SIGS Cross Research and Innovation Fund (JC2021005).}
\thanks{Xiankun Zhu, Yucheng Xin, Shoujie Li, Houde Liu and Bin Liang are with Shenzhen International Graduate School, Tsinghua University, Shenzhen 518055, China.}
\thanks{Chongkun Xia is with School of Advanced Manufacturing, Sun Yat-sen University, shenzhen 518107, China}
\thanks{{$\dagger$}Corresponding authors: Houde Liu (liu.hd@sz.tsinghua.edu.cn).}
}
\begin{document}

\maketitle
\thispagestyle{empty}
\pagestyle{empty}

\begin{abstract}

Fast and efficient collision detection is essential for motion generation in robotics. In this paper, we propose an efficient collision detection framework based on the Signed Distance Field (SDF) of robots, seamlessly integrated with a self-collision detection module. Firstly, we decompose the robot's SDF using forward kinematics and leverage multiple extremely lightweight networks in parallel to efficiently approximate the SDF. Moreover, we introduce support vector machines to integrate the self-collision detection module into the framework, which we refer to as the SDF-SC framework. Using statistical features, our approach unifies the representation of collision distance for both SDF and self-collision detection. During this process, we maintain and utilize the differentiable properties of the framework to optimize collision-free robot trajectories. Finally, we develop a reactive motion controller based on our framework, enabling real-time avoidance of multiple dynamic obstacles. While maintaining high accuracy, our framework achieves inference speeds up to five times faster than previous methods. Experimental results on the Franka robotic arm demonstrate the effectiveness of our approach. 
Project page: \href{https://sites.google.com/view/icra2025-sdfsc}{https://sites.google.com/view/icra2025-sdfsc}.

\end{abstract}

\section{INTRODUCTION}

Efficiently generating safe operational trajectories for robots is a critical challenge in the increasingly common context of human-robot interactionn\cite{c47}. Motion planning algorithms need to comprehensively consider external and self-collision issues. Much like how humans perceive obstacles, collision detection serves as the robot’s means of sensing environmental obstacles, ensuring it avoids collisions with its surroundings, humans, and itself \cite{c44}. Due to their low computational efficiency, traditional collision detection methods, which are generally based on the geometric information of obstacles and robot \cite{c1,c2}, struggle to meet the current demand for real-time generation of collision-free trajectories in complex and dynamic environments.

\begin{figure}[tbp]
\begin{minipage}[b]{.5\linewidth}
    \centering
    \subfigure[SDF and Self-Collision]{\label{SDF-SC}\includegraphics[width=0.85\linewidth]{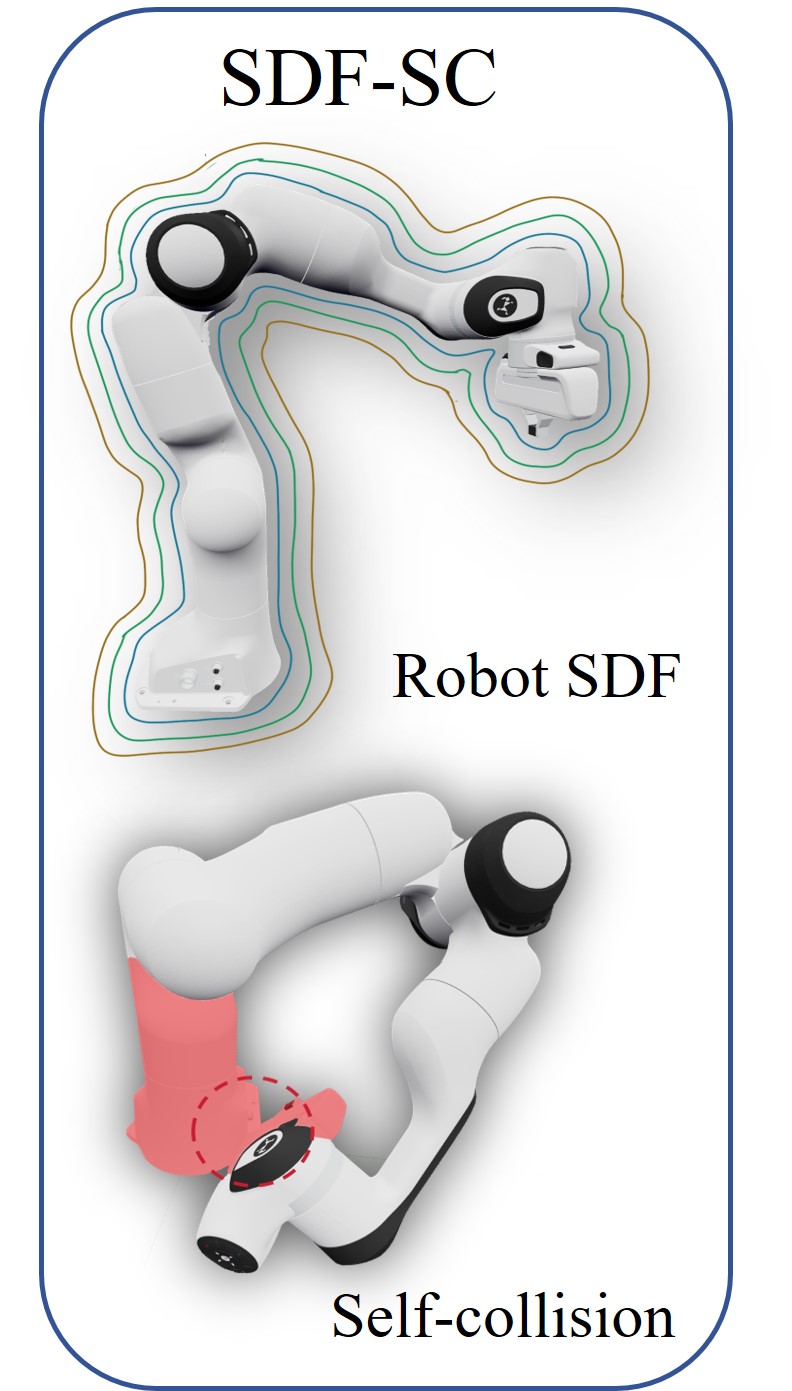}}
\end{minipage} 
\hfill 
\begin{minipage}[b]{.47\linewidth}
    \centering
    \subfigure[Trajectory Optimization]{\label{Trajectory Optimization}\includegraphics[width=.9\linewidth]{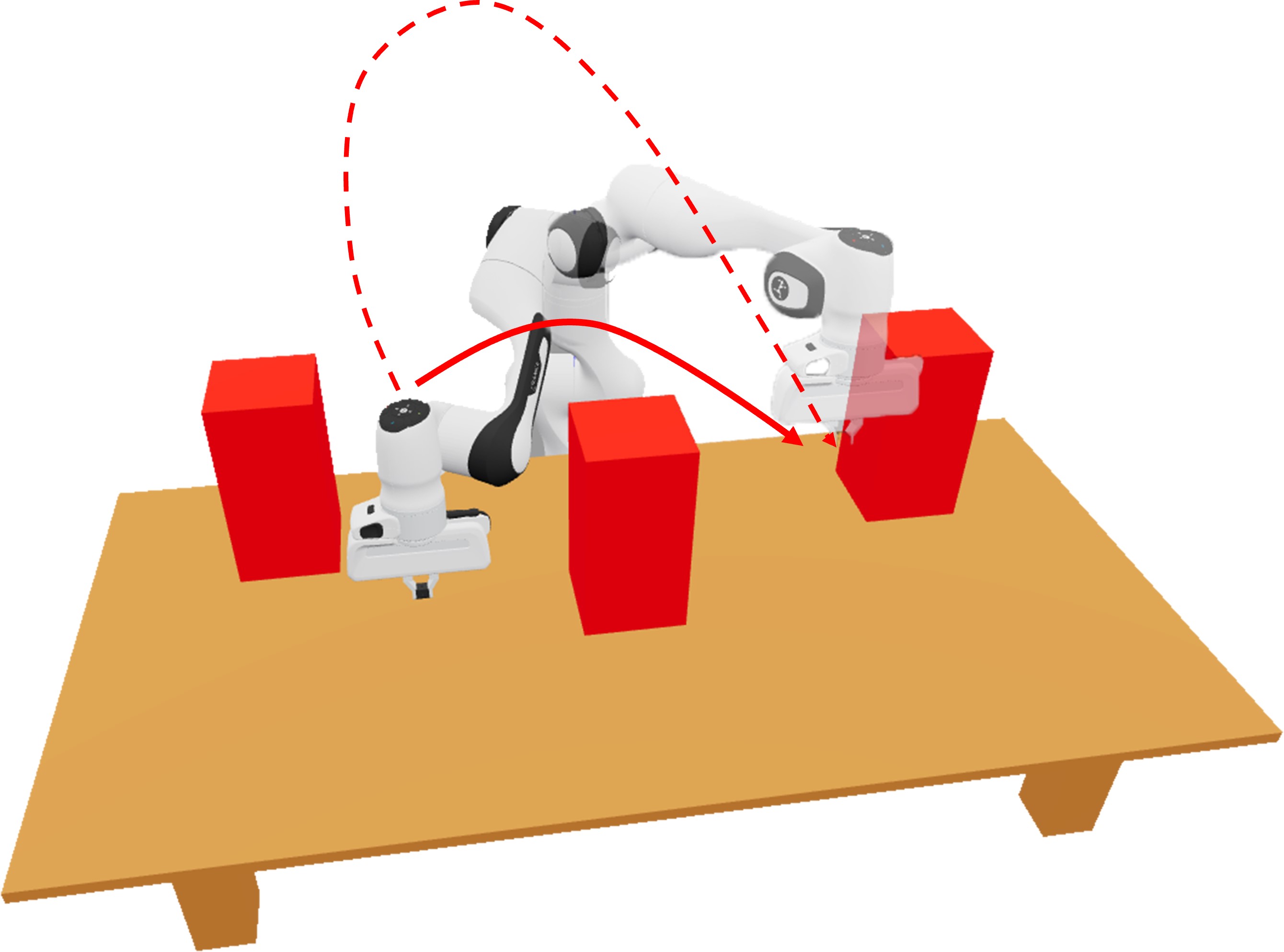}}
    
    \subfigure[Reactive Controller]{\label{Reactive Controller}\includegraphics[width=.9\linewidth]{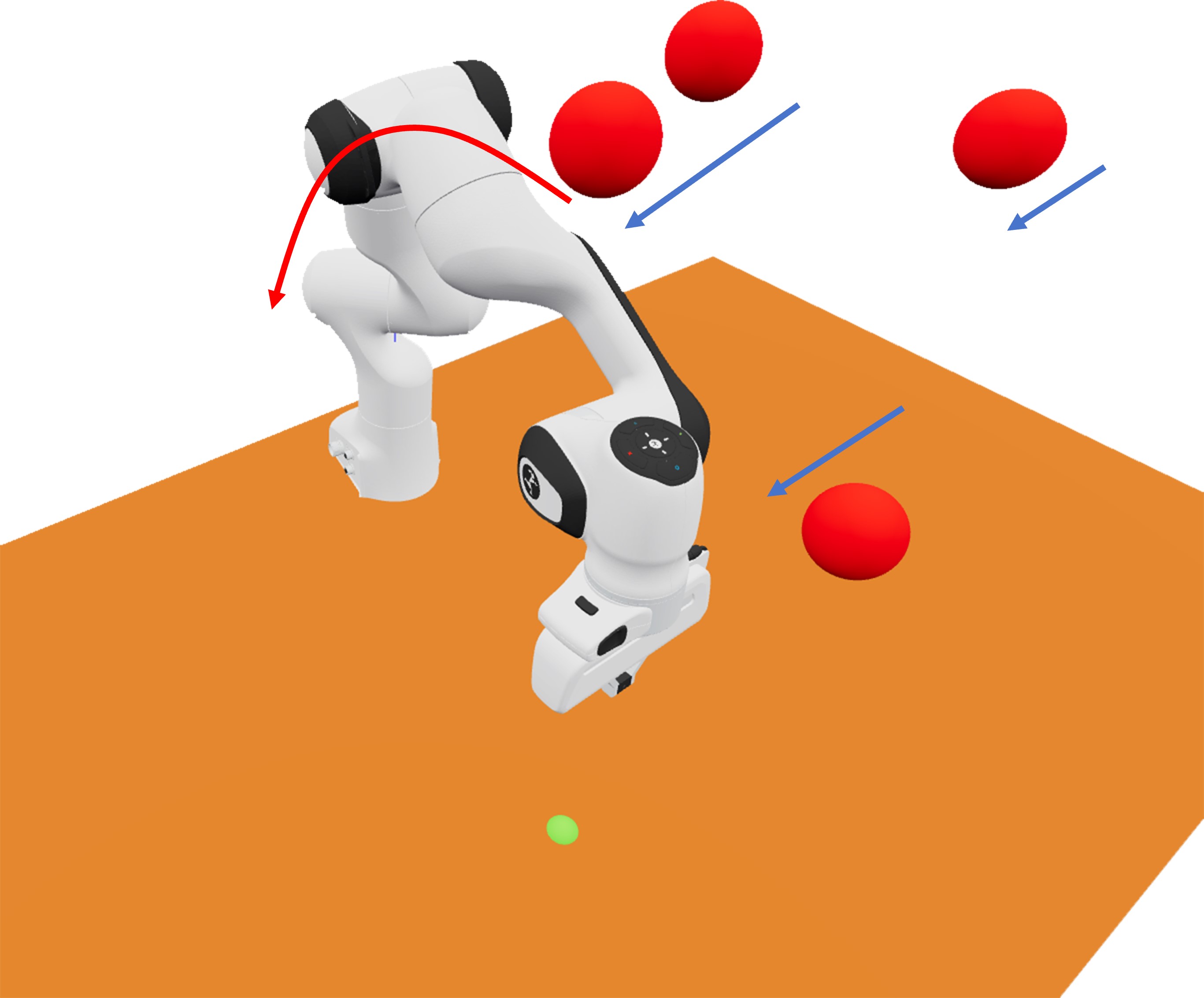}}
\end{minipage}
\vspace{-5pt}
\caption{Overview of our work. (a) We combine our own lightweight SDF framework with self-collision detection to obtain a new collision detection framework SDF-SC. (b) Trajectory Optimization based on the SDF-SC. (c) We use SDF-SC to enable reactive control of the robot.}
\label{fig:Overview}
\vspace{-15pt}
\end{figure}
The introduction of the Signed Distance Field (SDF) \cite{c6} concept offers a novel approach to addressing collision detection challenges. Originally a key research area in computer vision and graphics, SDF has been widely used to efficiently represent complex shapes \cite{c31,c43}. Migrating to the field of collision detection, SDF provides an intuitive and effective way of determining the shortest distance from a point to the surface of an object, which perfectly meets the collision detectors' demand for quantifying the collision distance between robots and obstacles. When applied to motion planning, SDF is extended to represent manifolds of general equality constraints \cite{c34}, facilitating real-time trajectory smoothing and enabling full-body control tasks\cite{c35,c36}. 
In learning SDF models, Koptev et al. \cite{c7} employ a multi-layer perceptron (MLP) to fit neural implicit signed distance functions. Liu et al. \cite{c11} propose a method that employs composite SDF networks to improve the fitting accuracy of articulated robots. Bernstein polynomials are employed as basis functions for SDF, further enhancing storage and computational efficiency \cite{c12}. These approaches learn SDF models coupled with different joint configurations. However, the complexity of network architectures or the verbosity of functional representations often limits the precision and speed of collision distance inference. Moreover, SDF has inherent limitations: it can only calculate the distance to external obstacles, but cannot account for self-collision, which is a critical and non-negligible issue in both simulations and the real world \cite{c27,c26}.


In this paper, we utilize a forward kinematic chain to optimize networks framework. During this process, we discover that extremely shallow networks are sufficient to achieve excellent learning outcomes for the SDFs of each link. Moreover, we employ a parallel framework to significantly accelerate inference, increasing the speed for thousands of points to five times that of previous methods \cite{c7,c11}. The joint boundaries of self-collisions and external collisions are robustly determined using a data-driven approach based on kernel perceptrons \cite{c9,c10}. Building on this, we represent the self-collision boundary using support vectors and develop a processing function that leverages the statistical properties of support vector machine (SVM) \cite{c20} results to calculate the self-collision distance.
By integrating self-collision and the external collision distance, we address a critical limitation of the prior SDF frameworks, which did not account for self-collisions during planning. This leads to the development of our final integrated system, termed the SDF-SC framework.

We apply the SDF-SC framework as inequality constraints within the optimization problem to evaluate its effectiveness in generating collision-free trajectories. Additionally, we integrate SDF-SC with a reactive control algorithm to tackle the challenge of a robot avoiding multiple dynamic obstacles. Experimental results demonstrate that this integration significantly enhances the robot's dynamic response to external and self-collision risks, leading to more robust and adaptable performance in real-world scenarios.



The contributions of our work are as follows:

\begin{itemize}
    \item We propose an efficient and lightweight parallel networks framework for learning SDF, achieving high fitting accuracy while significantly accelerating inference speed.
    \item We augment the framework by incorporating integrated self-collision avoidance, while maintaining continuity and differentiability throughout the process.
    \item We demonstrate the effectiveness of SDF-SC framework through rigorous trajectory optimization experiments and leverage it to design a robust reactive controller.
\end{itemize}

\section{Framework Composition}

\begin{figure}
    \centering
    \includegraphics[width=1.0\linewidth]{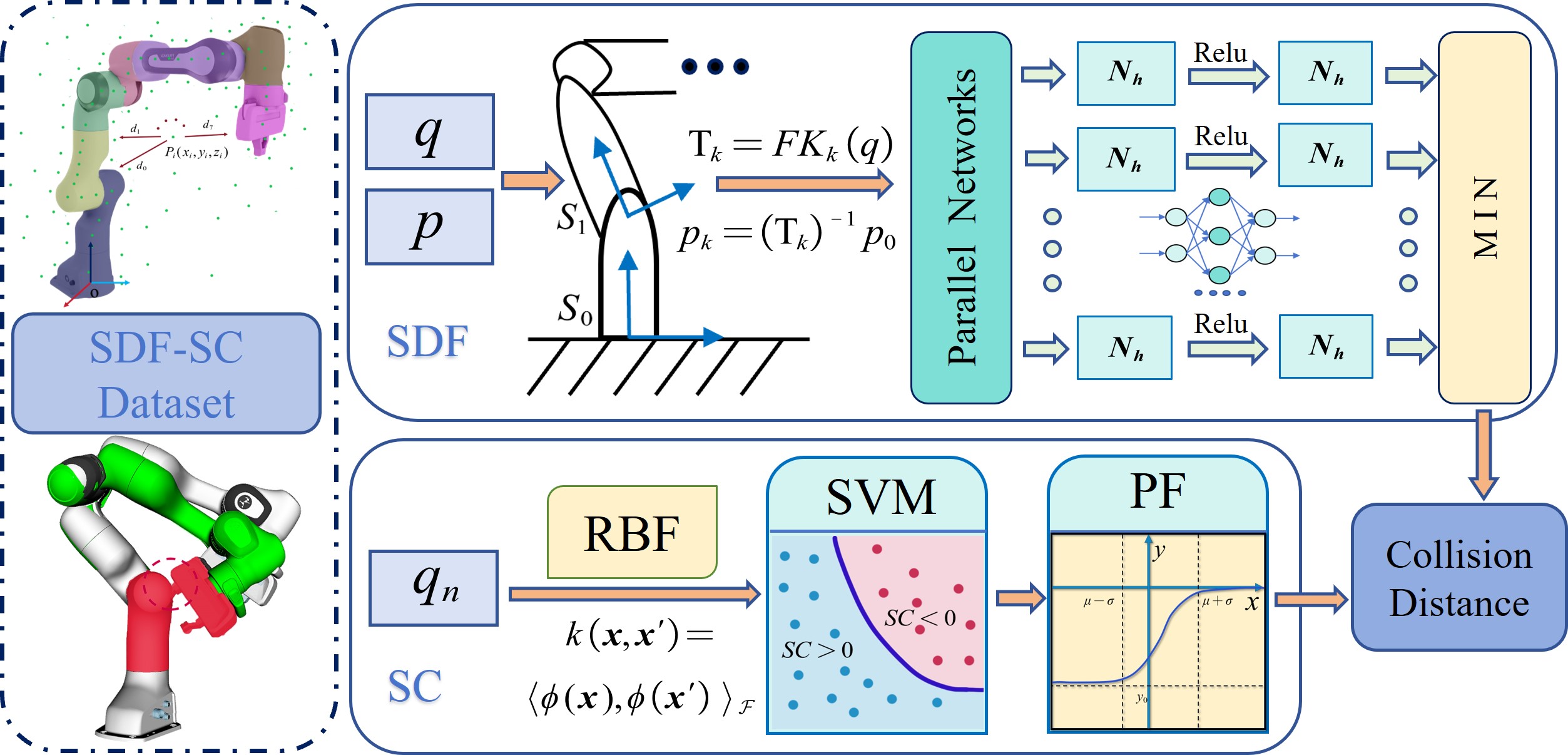}
    \caption{Overall algorithm pipeline for estimating collision distance. Collision distance refers to the safety margin between the robot and potential collisions.}
    \label{fig:Overall algorithm}
    \vspace{-10pt}
\end{figure}

In this section, we introduce a holistic framework for integrated collision detection, which includes a parallel networks framework for learning the SDF of articulated robots, an algorithm for self-collision evaluation, and dataset construction. The overall algorithmic pipeline for estimating the collision distance is depicted in Fig. \ref{fig:Overall algorithm}, with the Franka Emika Panda robot serving as the subject for demonstration.
\subsection{Parallel Signed Distance Field Networks} 
Consider an articulated robot with $n$ degrees of freedom and $K$ links, where the joint angles are defined as $\boldsymbol{q}=$ $\left[q_1, q_2, \ldots, q_n\right]^{\top} \in \mathbb{R}^n$. Each link is considered as a rigid body, and the coordinate system $\left\{\mathbf{S}_k\right\}_{k=1}^K$  is rigidly attached to the rigid body. Let $\mathbf{S}_0$ denote the base coordinate system, and $\mathbf{T}_k\in \mathbb{S} \mathbb{E}(3)$ represent the homogeneous transformation matrix from $\mathbf{S}_0$ to $\mathbf{S}_k$.
$$
\mathbf{T}_k=FK_k\left( \boldsymbol{q} \right)  \eqno{(1)}
$$
where $\mathbf{T}_k$ is determined by the Denavit-Hartenberg (DH) parameters and joint angles $\boldsymbol{q}$ of the articulated robot, and the computation of the transformation matrix can be achieved through the forward kinematic chain. We map 3D query 
 points $\boldsymbol{p} \in \mathbb{R}^3$ from the base coordinate system onto the coordinate systems of each link through matrix transformation, which can be written as
$$
\boldsymbol{p}_k=\left(\mathbf{T}_k\right)^{-1} \boldsymbol{p} \eqno{(2)}
$$
where $\boldsymbol{p}_k$ denotes the coordinate representation of point $\boldsymbol{p}$ in the coordinate system of each link. Owing to the rigid attachment of the coordinate system to the link, the metric distance  $\boldsymbol{d}_k$ from a point to the 
$k^{t h}$ link is solely determined by $\boldsymbol{p}_k$. It should be noted that when a point is located within the robot, we represent its distance to the surface of the links with negative values.  

The distance from a point to an entire articulated robotic system is defined as the minimum distance from that point to each individual link within the system. For convenience, we employ the function 
$\Gamma(\boldsymbol{q}, \boldsymbol{p})$ to represent the SDF value from a point $\boldsymbol{p}$ to the articulated robot with joint configuration $\boldsymbol{q}$, with the following equation:
$$
\Gamma(\boldsymbol{q}, \boldsymbol{p})=\min (\boldsymbol{d}_1(\boldsymbol{p}_1),
\boldsymbol{d}_2(\boldsymbol{p}_2),
\cdots,
\boldsymbol{d}_K(\boldsymbol{p}_K))
\eqno{(3)}
$$
Deviating from the traditional approach of directly mapping the relationship from the joint configuration $\boldsymbol{q}$
and point $\boldsymbol{p}$ to the fuction $\Gamma(\boldsymbol{q}, \boldsymbol{p})$ via a network \cite{c7,c17}, our methodology employs the learning of the distance functions $\boldsymbol{d}_k$ for each link to effectively fit $\Gamma(\boldsymbol{q}, \boldsymbol{p})$.

 We employ a series of lightweight neural networks to approximate $\boldsymbol{d}_k(\boldsymbol{p}_k)$, with paralleling these networks to expedite the computation speed. The minimum output among them is defined as SDF value for the whole articulated robot.

\subsection{Self-Collision Detection}
Calculating the geometric distances associated with self-collisions in robots is an exceedingly laborious and time-consuming process\cite{c18}. However, due to the static and unique nature of self-collision regions in the joint space of articulated robots \cite{c40,c49}, it is feasible to learn a continuous differentiable function that models the self-collision boundary between colliding and free joint configurations.

To learn the self-collision function, we
perform uniform sampling of configurations within the joint space and subsequently categorize them into binary classes based on whether they collide or not. The sampled configurations $q_i$ are constrained within the joint limits $\boldsymbol{q}^{-},\boldsymbol{q}^{+}$ , and these constraints are utilized to normalize all configurations to $q_n$:
$$
\boldsymbol{q}^{-}<q_i<\boldsymbol{q}^{+}, i=1, \ldots, N_s \eqno{(4)}
$$
The label $y_i=+1$ is used to denote that $q_i$ is in the self-collision-free class, while $y_i=-1$ indicates that $q_i$ is in the self-colliding class.
Subsequently, we employ a SVM algorithm to construct a model for self-collision detection, where the decision function takes the following form:
$$
S(\boldsymbol{q})=\sum_{i=1}^{N_{v}} \alpha_i y_i K\left(\boldsymbol{q}, \boldsymbol{q}_i\right)+b \eqno{(5)}
$$
where $\boldsymbol{q}_i$ and $y_i$ represent the support vectors obtained from the SVM algorithm and their respective collision labels. $N_{v}$ is the number of support vectors. $0 \leq \alpha_i \leq C$ denotes the weight of the support vectors and $C$ is an adjustable parameter that serves as a penalty for errors, helping to prevent overfitting.
$b \in \mathbb{R}$ denotes the bias term in the decision rule.

Within the purview of kernel function selection, empirical evidence from prior studies has demonstrated that both the Radial Basis Function (RBF)  kernel and forward kinematics (FK) kernel \cite{c21} manifest commendable efficacy. However, in deference to the exigencies of rapid self-collision detection, the RBF kernel $K\left(\boldsymbol{q}, \boldsymbol{q}_i\right)=e^{-\gamma\left\|\boldsymbol{q}-\boldsymbol{q}_i\right\|^2}$ has been selected for its computational efficiency.

\subsection{Dataset Construction}
In order to facilitate algorithmic comparison with prior work, it is imperative to ensure the comparability of our dataset with those utilized previously. Consequently, our sampling methodology aligns with the approaches employed in prior studies for dataset construction \cite{c7,c11}, where data points $\boldsymbol{p}$ are more densely collected near link surfaces. Additionally, we have supplemented our dataset with distances obtained from the GJK in simulation environments,  providing a valid supplement. 
Ultimately, the dataset for each link comprises 50 million entries.

In preparation for self-collision detection, we create training and validation datasets containing 100,000 and 30,000 configurations, along with a test set of 50,000 configurations. Self-collision labels are generated using the self-collision detection function from the FCL library \cite{c1}. 
\section{Hyperparameter Search and  Framework Evaluation}
\subsection{Hyperparameter Search}
\begin{figure}[t]
\begin{minipage}[b]{.5\linewidth}
    \centering
    \subfigure[]{\label{fig:SVM score}\includegraphics[width=1\linewidth]{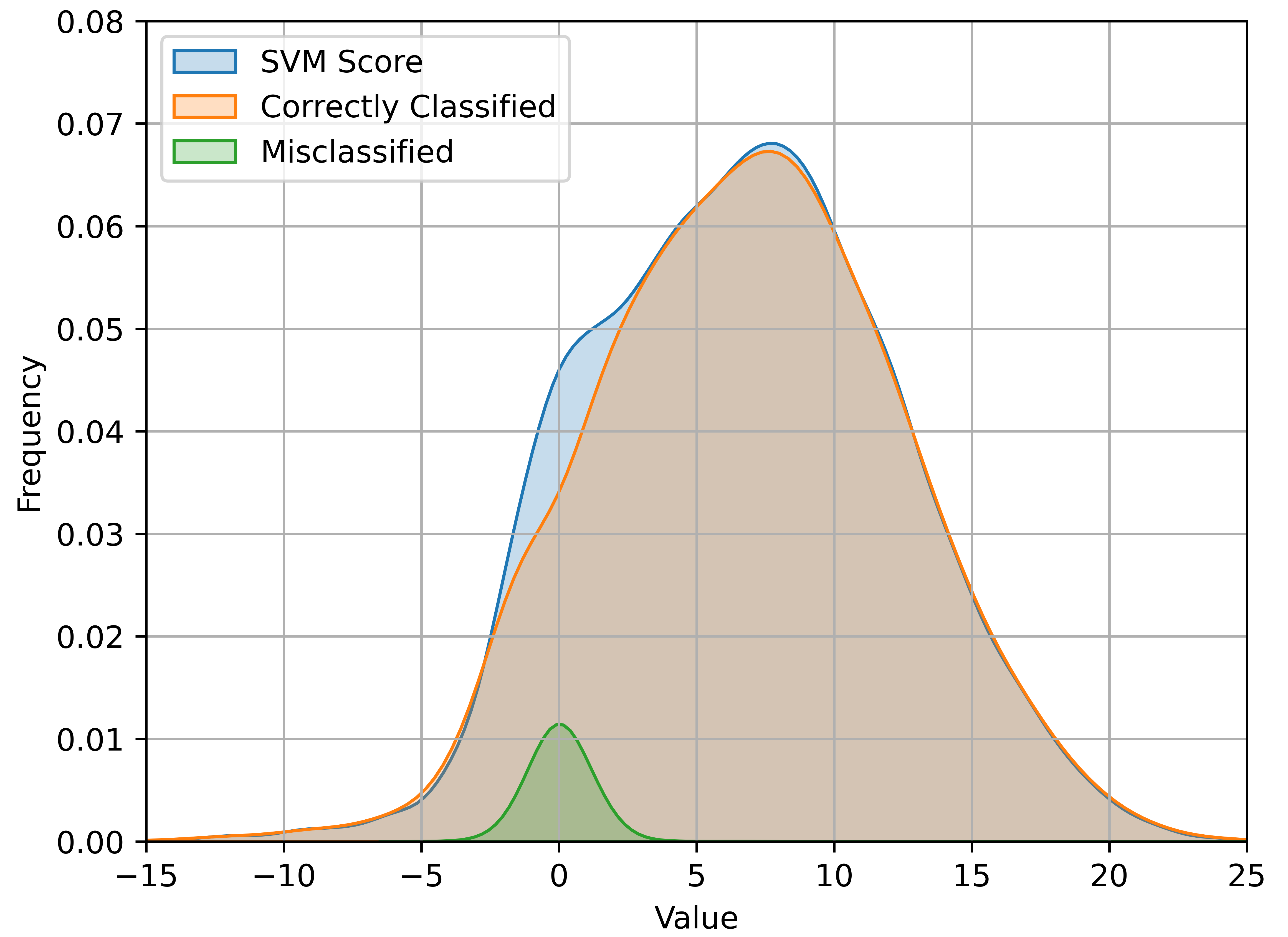}}
\end{minipage} 
\hfill 
\begin{minipage}[b]{.49\linewidth}
    \centering
    \subfigure[]{\label{fig:nn-loss}\includegraphics[width=1\linewidth]{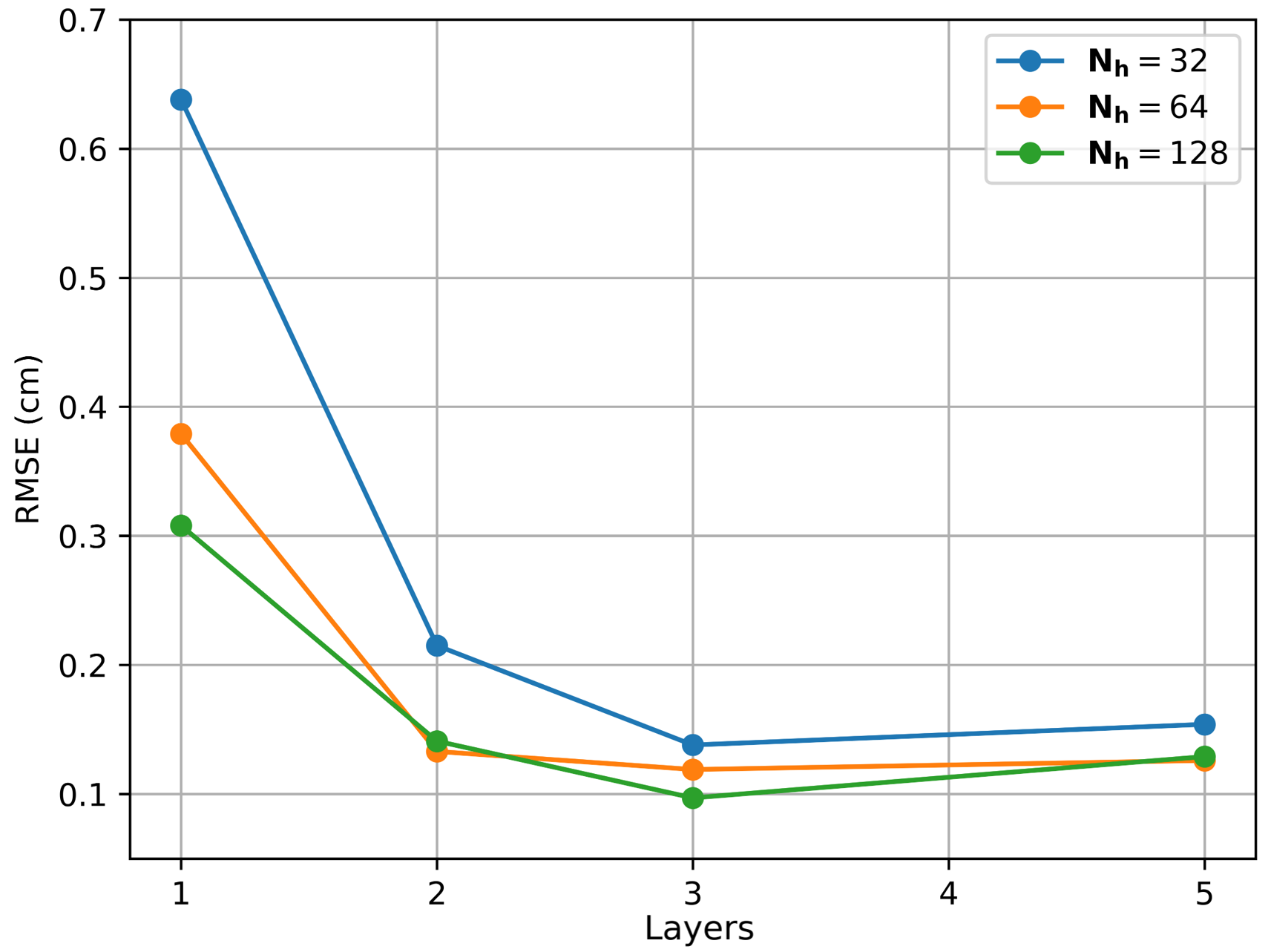}}
\end{minipage}
\vspace{-5pt}
\caption{(a) The score distribution of the self-collision detection model on the validation set. (b) The impact of different network architectures on fitting accuracy.}
\label{fig:all}
\vspace{-12pt}
\end{figure}

Following hyperparameter optimization of the SVM model, we select $C=50$ and 
$\gamma=1.0$ as the optimal parameters, resulting in a total of $N_{v}=10446$ support vectors that constitute the self-collision boundary. The score distribution of our trained model, as exhibited on the validation set, is depicted in Fig. \ref{fig:SVM score},  indicates that $S(\boldsymbol{q})=0$ represents the self-collision boundary constituted by the support vectors. When configurations are classified for self-collision based on the boundary, the distribution of misclassified configuration scores approximates a Gaussian distribution $X \sim N\left(\mu, \sigma^2\right)$.

For the seamless integration and enhancement of self-collision detection with collision distance metrics, the robotic configurations are delineated into three distinct regions based on statistical features: distant, proximal, and transgressing the self-collision boundary. The proximal region of the self-collision boundary is delineated according to Gaussian distribution principles as the interval $(u-\sigma, u+\sigma)$.

Within our framework, the collision distance is computed using the following formula:
$$
D(\boldsymbol{q}, \boldsymbol{p})=\Gamma(\boldsymbol{q}, \boldsymbol{p})+P(S(\boldsymbol{q})) \eqno{(6)}
$$
where $P$ denotes a processing function for the self-collision scores. The equation for $\partial D$ is naturally derived as follows:
$$
\frac{\partial D(\boldsymbol{q}, \boldsymbol{p})}{\partial \boldsymbol{q}}=\frac{\partial \Gamma(\boldsymbol{q}, \boldsymbol{p})}{\partial \boldsymbol{q}}+\frac{\partial P}{\partial S} \cdot \frac{\partial S}{\partial \boldsymbol{q}} \eqno{(7)}
$$
$$
\frac{\partial S}{\partial \boldsymbol{q}}=-2 \gamma \sum_{i=1}^{N_{v}} \alpha_i y_i e^{-\gamma\left\|\boldsymbol{q}-\boldsymbol{q}_i\right\|^2}\left(\boldsymbol{q}-\boldsymbol{q}_i\right) \eqno{(8)}
$$

In the case of single-arm robots, such as the Franka, self-collision configurations represent a special case, with their occurrence being significantly less frequent than external collisions. Consequently, we expect the collision distance to approximate the robot's SDF in regions deep within the self-collision boundary, and to invert to negative values as an indicator of constraint when configurations transgress the boundary. Furthermore, the gradient 
$\partial P$ is aligned with $\partial \Gamma$, both directing towards increasing collision distance.
\begin{table}[t]
\caption{Performance metrics of the self-collision detection}
\vspace{-7pt}
\label{SC}
\begin{center}
\begin{tabular}{c|c|c|c}
\hline
Dataset & Acc & TPR & TNR\\
\hline
Train   &  0.99 &0.95 & 0.99\\
Validation & 0.97 & 0.86 & 0.98\\
Test & 0.97 & 0.88 & 0.98\\
\hline
\end{tabular}
\end{center}

\end{table}

To this end, we formulate a continuous, monotonically increasing processing function $P$, bounded both above and below, as presented below:
$$
P(s)=-\frac{1}{1+e^{k s+b_0}} \eqno{(9)}
$$
where the parameters $k$ and $b_0$ are determined by setting a suitable interval for $P(u-\sigma, u+\sigma)$ in the range $[-0.95,-0.001]$ to fit the statistical characteristics of $S$.

Before evaluating the performance of external collision distance detection, we conduct an optimization search for the number of layers and nodes in the parallel SDF network to determine the most effective architectural parameters for the networks. The layer count ranges from 1 to 5, while the node count varies between 32, 64, and 128. Each network in the parallel configuration uniformly employs a Root Mean Squared Error (RMSE) loss function, with the Adam optimizer used consistently across all. The ReLU function is selected as the activation function and the maximum number of iterations is set to 100 to ensure network convergence.

The impact of different network frameworks on the final RMSE convergence is illustrated in Fig. \ref{fig:nn-loss}. Experiments reveal that a parallel configuration comprising merely two layers of networks is sufficient to achieve an excellent fit for the articulated robot's SDF, with additional layers providing minimal benefit in terms of RMSE reduction and, conversely, increasing the inference time. To balance accuracy and inference speed, we select a configuration of 2 layers with 64 nodes as the final architecture parameters for our parallel SDF networks, which makes our model significantly  lighter compared to previous work \cite{c11,c7}.
\subsection{Framework Evaluation}
All experiments are conducted using an Intel Core i7-13700KF and an NVIDIA GeForce RTX 4070 GPU, with all GPU operations performed using batch processing.

To evaluate the framework's effectiveness for self-collision detection, we evaluate it on the test dataset, employing Accuracy (Acc), True Negative Rate (TNR), and True Positive Rate (TPR) as metrics. The collision distance $D(\boldsymbol{q}, \boldsymbol{p})$ boundary for classification is set at $u-\sigma$. A high TPR prevents the misclassification of collision-free configurations as colliding, while a low TPR may result in overly conservative detection. On the other hand, a high TNR is the most critical metric that should be maximized, as it prevents the erroneous labeling of colliding configurations as free. The self-collision detection model within our framework achieves commendable performance across various datasets, maintaining a TNR of over 0.98 in all cases, as detailed in the Table \ref{SC}.

In our experimental analysis, we evaluate the mean computational latency of our comprehensive framework, denoted as SDF-SC, across a spectrum of 3D query points. This evaluation is conducted in comparison with the relevant Composite-SDF \cite{c11}, Neural-JSDF \cite{c7}, RDF \cite{c12} and the standard Gilbert-Johnson-Keerthi (GJK) \cite{c2} algorithm. The computations within the SDF-SC framework encompass not only the forward kinematics chain and the parallel SDF networks but also self-collision detection. 
\begin{table}[t]
\caption{Comparison of computational times for collision distance by various algorithms}
\label{SDF time}
\begin{center}
\begin{threeparttable}
    \begin{tabular}{c|c|c|c|c}
    \hline
    Number of Query Pointss & 1 & 100 & 1000 & 10000 \\
    \hline
    SDF-SC \textbf{(ours)} ,ms & 0.98 & 1.02 & \textbf{1.05} & \textbf{1.21} \\
    Composite SDF\tnote{1} ,ms & 4.33 & 4.78 & - & 6.63 \\
    Neural-JSDF ,ms   & \textbf{0.18} & \textbf{0.42} & 1.13 & 8.98 \\
    RDF(N=8) ,ms  & 1.12 & 1.21 & 1.27 & 2.12 \\
    GJK ,ms           & 0.95 & 87.4 & 815  & 7984\\
    \hline
    \end{tabular}
    \begin{tablenotes}   
        \footnotesize              
        \item[1] The data cited in this row is derived from paper \cite{c11}.       
    \end{tablenotes}           
\end{threeparttable} 
\end{center}
\end{table}
\begin{figure}[t]
    \centering
    \includegraphics[width=1\linewidth]{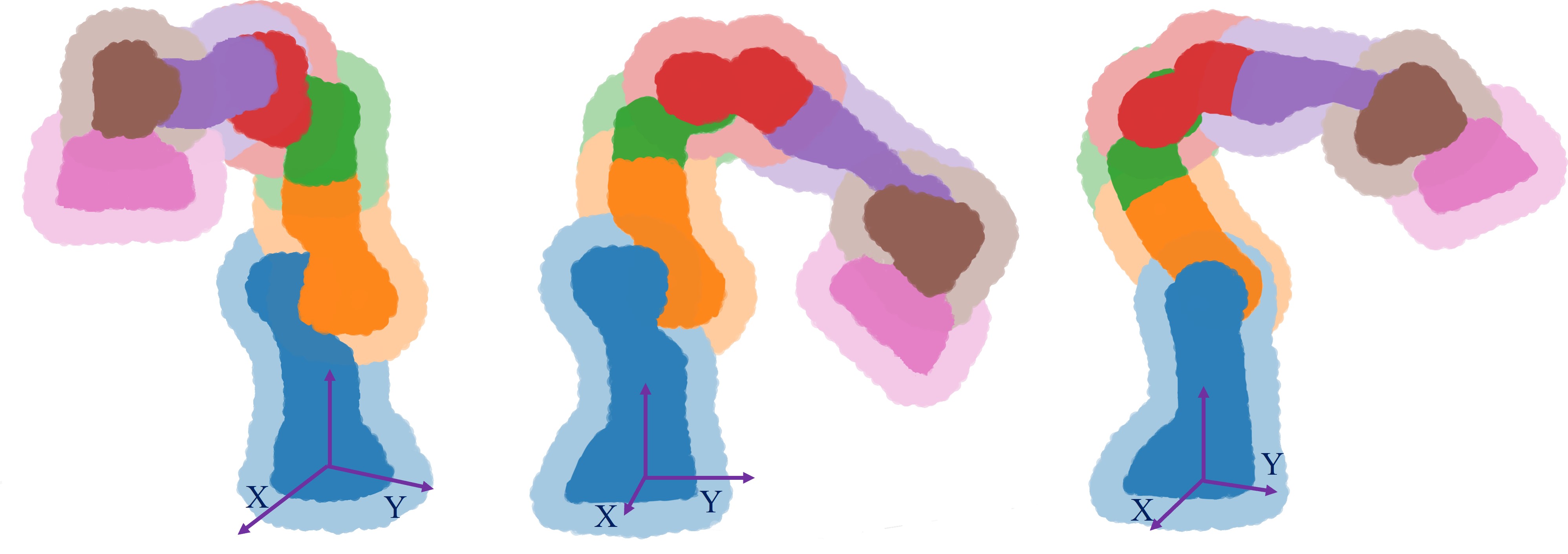}
    \caption{The reconstruction of the Franka Emika Panda's links is performed based on the distance isosurfaces $D(\boldsymbol{q}, \boldsymbol{p})=$0 cm (solid) and 
$D(\boldsymbol{q}, \boldsymbol{p})=$5 cm  
 (transparent). Different colors represent different links.}
    \label{fig:FRANKA-SDF}
\vspace{-15pt}
\end{figure}
The experimental results are presented in Table \ref{SDF time}. Despite the comprehensive nature of these computations, our detection efficiency remains highly superior, exhibiting only a slight disadvantage when the number of query points is relatively low in comparison to the Neural-JSDF algorithm. However, in general scenarios where environmental modeling often requires sampling several thousand points, our detection efficiency is generally five times greater than that of previous methods in standard applications.

The average RMSE across the links of the articulated robot serves as a metric for the quality of SDF fitting. Distances are categorized with 
$\boldsymbol{d}<$0.1 m as the region close to the robot and 0.1 m$<\boldsymbol{d}<$1.2 m as the region distant from the robot. We compare our results within these two regions with those of the Neural-JSDF, Composite SDF and RDF algorithms. The results are as shown in the Table \ref{tab:SDF accuracy }. 
\begin{table}[h]
\vspace{-5pt}
\caption{Comparing the SDF accuracy obtained by the algorithms}
    \centering
    \begin{tabular}{c|c|c}
    \hline
        Region & $\boldsymbol{d}<$ 0.1 m & 0.1 m $<\boldsymbol{d}<$ 1.2 m  \\
        \hline
        SDF-SC \textbf{(ours)}, cm & \textbf{0.16} & \textbf{0.28}\\

       Composite SDF, cm  & 0.21 & 0.36\\ 
       
       Neural-JSDF, cm & 1.04 & 1.06 \\
       RDF(N=8), cm  & 0.41 & 0.54 \\
       \hline
    \end{tabular}
    \label{tab:SDF accuracy }
\vspace{-5pt}
\end{table}

Our method exhibits superior accuracy in both distance intervals, with an overall precision of 0.19 cm, attaining millimeter-level accuracy. To more vividly demonstrate the accuracy of our model, we utilize the integrated framework to generate a 3D point cloud, thereby reconstructing the surface of the Panda robot at $D(\boldsymbol{q}, \boldsymbol{p})=$0 cm. As illustrated in Fig. \ref{fig:FRANKA-SDF}, the SDF reconstructed by our method accurately reflects the surface contours of the Panda robot, with very uniform widths between adjacent isosurfaces, indicating that the integrated framework can effectively obtain the precise SDF of the articulated robot.

The evaluations conducted indicate that the integration of self-collision and external collision detection into a unified collision distance $D(\boldsymbol{q}, \boldsymbol{p})$ within our framework is viable, and it performs effectively in detecting both types of collisions.

\section{Montion Planning Experiments}
In this section, we substantiate the superiority of our framework in robotic motion planning through two categories of experiments: 1) \textit{Trajectory Optimization} 2) \textit{Reactive Control}. During planning, we utilize 
$D(\boldsymbol{q}, \boldsymbol{p})$ as the collision distance to simultaneously avoid self-collisions and external collisions.
\subsection{Trajectory Optimization with SDF-SC}

To achieve the most efficient trajectory while meeting the requirement of avoiding static obstacles, we define the optimization problem as follows:
$$
\min _q \quad f(  \boldsymbol {q})=\sum_{t=0}^{T-1} \|FK\left(\boldsymbol{q}_{t+1}\right) -FK\left(\boldsymbol{q}_t\right)\|^2 
$$
$$
\begin{aligned}
\hspace{-1.8cm}\text {subject to } \quad \quad \quad 
D(\boldsymbol{q},\boldsymbol{p})+\epsilon  &\leq 0
\\
\|{\boldsymbol{q}_{t+1}-\boldsymbol{q}_{t}}\| &\leq \boldsymbol{w}_{\max }
\\
 \|\mathbf{e e}(\boldsymbol{q}_{t+1})-\mathbf{e e}(\boldsymbol{q}_{t})\| &\leq  \boldsymbol{v}_{\max }
\\
\boldsymbol{q}^{-} \leq \boldsymbol{q} &\leq \boldsymbol{q}^{+} 
\end{aligned}
\eqno{(10)}
$$
where $f(\boldsymbol {q})$ denotes the cost function of the optimization problem. $T$ represents the number of planned path points or time duration and $FK$ stands for the forward kinematics chain of each link. $\epsilon$ denotes the safety margin, which is employed to augment the level of conservatism in collision avoidance. $\mathbf{e e}$ denotes the position of the end-effector, while $\boldsymbol{v}_{\max }$ represents the maximum velocity for the end-effector.

\begin{figure}[t]
    \centering
    \includegraphics[width=0.9\linewidth]{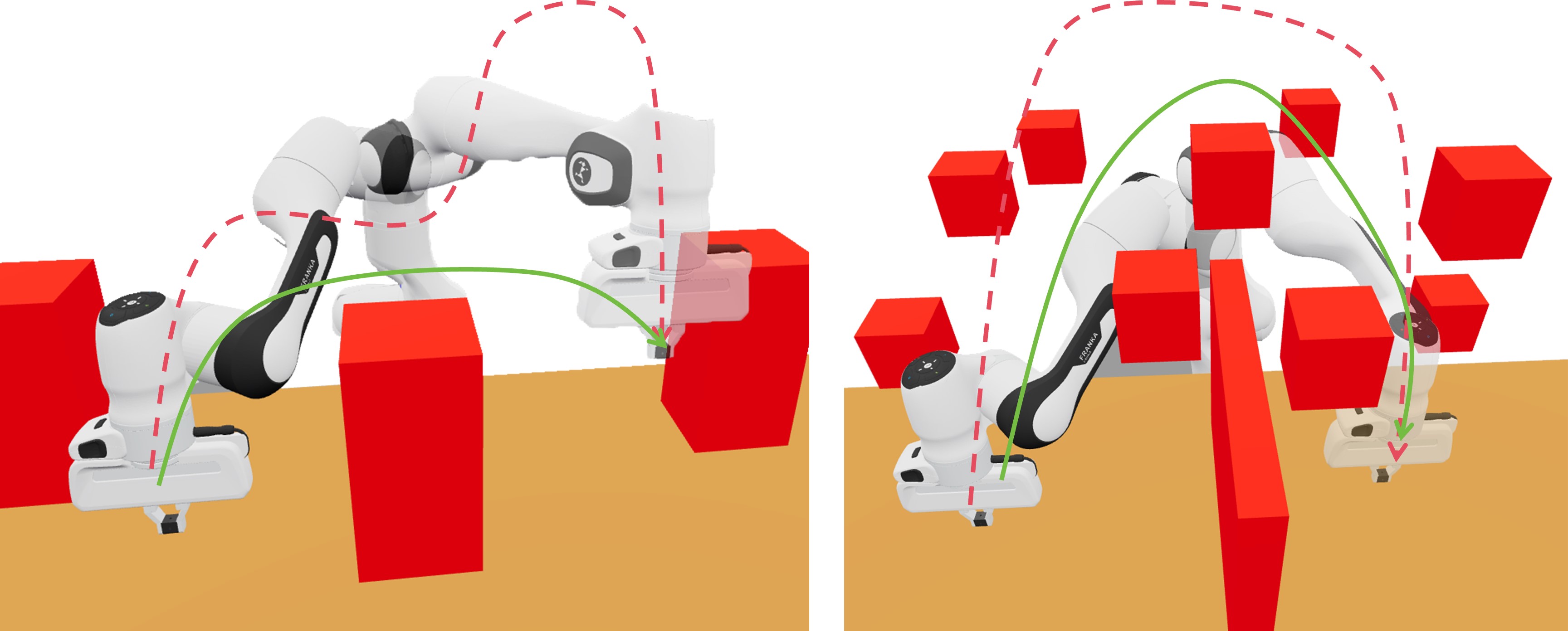}
    \caption{Trajectory optimization in simple (left) and complex (right) scenarios, where the red dashed line represents the trajectory generated by RRT*, and the green dashed line represents the trajectory optimized using SDF-SS with RRT-connect.}
    \label{fig:traj}
\end{figure}

\begin{figure}[tbp]
\begin{minipage}[b]{.45\linewidth}
    \centering
    \subfigure[]{\label{fig:traj-dist}\includegraphics[width=1\linewidth]{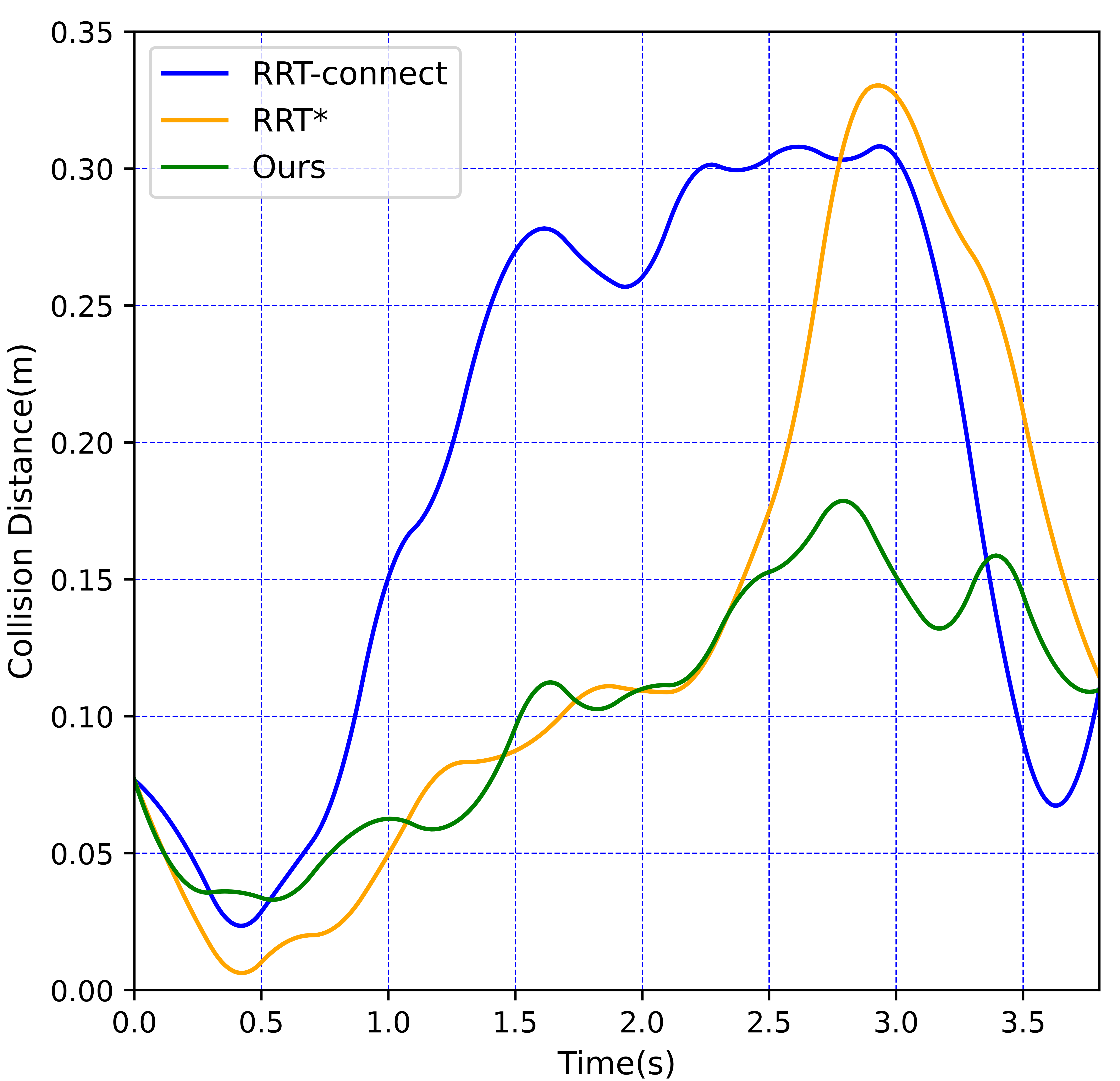}}
\end{minipage} 
\hfill 
\begin{minipage}[b]{.53\linewidth}
    \centering
    \subfigure[]{\label{fig:Comparative analysis}\includegraphics[width=1\linewidth]{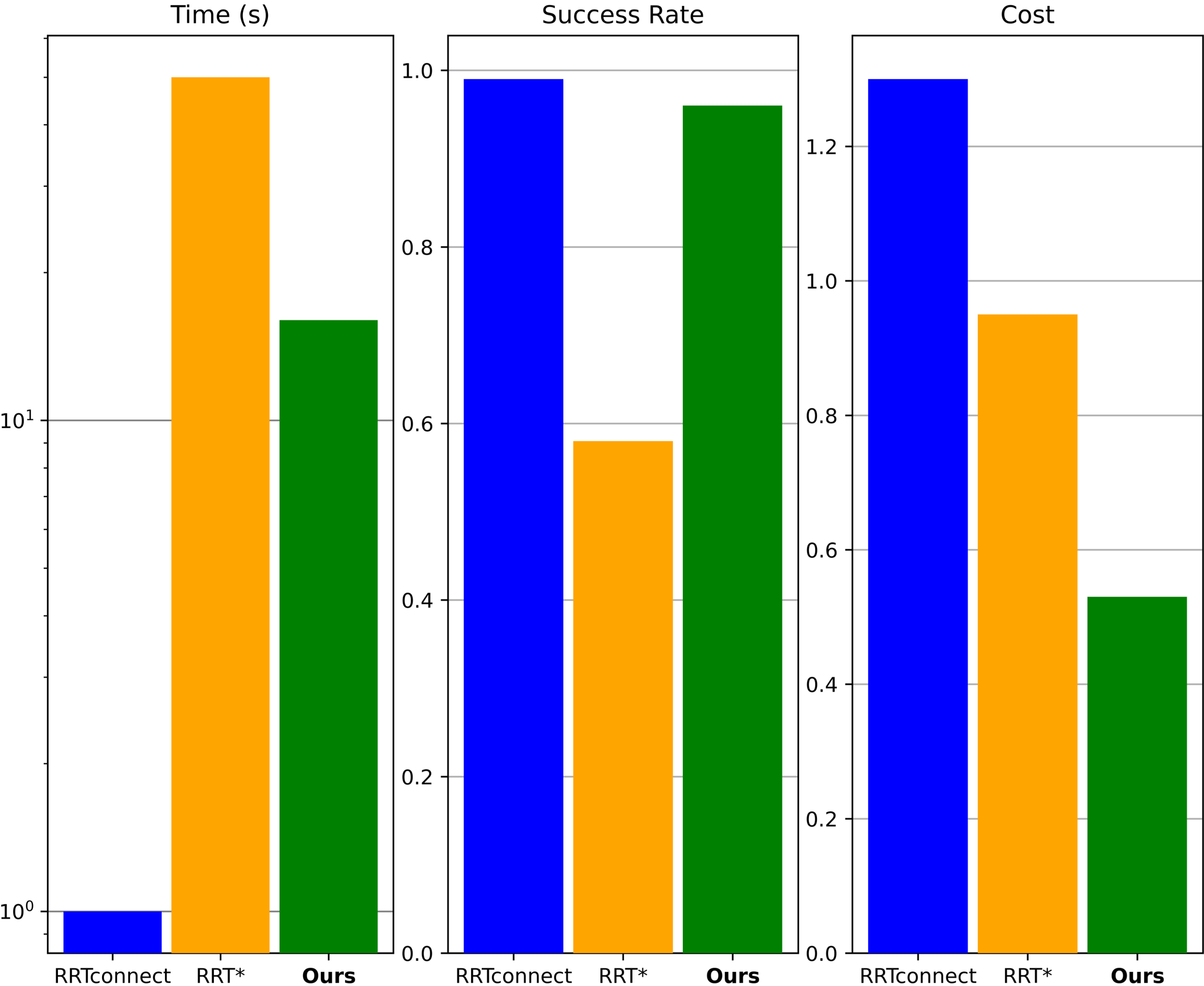}}
\end{minipage}
\vspace{-5pt}
\caption{(a) Collision distance during the movement period. (b) Comparative analysis of trajectory planning methods.}
\label{fig:all2}
\vspace{-12pt}
\end{figure}

To demonstrate the broad effectiveness of our optimization algorithm, we design scenarios with a simple environment containing three obstacles and a complex environment containing ten obstacles, as illustrated in Fig. \ref{fig:traj}. In our approach, we utilize the RRTconnect \cite{c13} algorithm to swiftly generate an initial trajectory, followed by employing the SLSQP \cite{c41} optimizer for our optimization problem, thereby obtaining the final optimized trajectory. Since our framework is differentiable, the gradients of the constraint functions can be obtained using the automatic differentiation functions in the PyTorch library. As a benchmark, we use the RRT* algorithm\cite{c30}, which is widely applied in path planning and theoretically proven to achieve asymptotic optimality \cite{c5}. The maximum iteration time for RRT* is set to 50 s. Fig. \ref{fig:traj-dist} shows the variation of collision distance over the movement time during our trajectory planning experiments in a simple scenario for various algorithms. By minimizing the path cost, our approach allows the robot arm's trajectory to approach obstacles more closely, without exceeding safety margins. 

Additionally, we perform 50 planning experiments in both simple and complex scenarios, and then take the average of the results from both sets. The comparative results are illustrated in Fig. \ref{fig:Comparative analysis}. Benefiting from the fully differentiable nature of our framework, which requires only approximately 3 ms per differentiation, we significantly reduce the path cost in trajectory planning using SDF-SC significantly, while maintaining a high success rate and relatively fast computation time. 

To validate the self-collision detection capabilities of our framework, we deliberately introduce self-collision configurations into the initial trajectories. Our framework is tasked with detecting the presence of self-collisions along these trajectories and optimizing them to eliminate such incidents. We design approximately 20 such trajectories, and our system successfully detects 95\% of the self-collisions, while also optimizing the trajectories to avoid these points. However, it is important to note that due to the use of a steep gradient to escape self-collisions, the trajectory transitions were not as smooth as desired, necessitating additional interpolation to refine the affected segments. Our framework is capable of detecting both external and self-collisions within the trajectory,  while also using gradient guidance to optimize trajectory and avoid collisions.


\subsection{Reactive Control with SDF-SC}
Inspired by NEO\cite{c16}, we utilize the SDF-SC framework to design a reactive controller, designated as NEO-SS, intended for real-time dynamic obstacle avoidance. We define the following quadratic programming (QP) problem \cite{c29} for the controller:
$$
 \min _x \quad g(\boldsymbol{x})=\frac{1}{2} \boldsymbol{x}^{\top} \mathcal{Q} \boldsymbol{x}+\mathcal{C}^{\top} \boldsymbol{x} 
$$
$$
\begin{aligned}
\hspace{-1.8cm} \text { subject to }  \quad
 \mathcal{J} \boldsymbol{x}&=\boldsymbol{\nu} \\
 \mathcal{A} \boldsymbol{x} &\leq \mathcal{B} \\
 \mathcal{X}^{-} &\leq \boldsymbol{x} \leq \mathcal{X}^{+}
\end{aligned}
\eqno{(11)}
$$
where 
$$
\boldsymbol{x}=\binom{\dot{\boldsymbol{q}}}{\boldsymbol{\delta}} \in \mathbb{R}^{(n+6)} \eqno{(12)}
$$
$\boldsymbol{\delta}$ is a slack variable with respect to 
$\dot{\boldsymbol{q}}$, representing the difference between the desired and actual end-effector velocity $\boldsymbol{\nu}$. $\mathcal{C}$ maximizes the manipulability of the robot, which is synthesized from $
\boldsymbol{J}_m$ \cite{c45} and the identity matrix. In NEO, the inequality constraints of a QP problem are represented as:
$$
\boldsymbol{J}_{d_l}(\tilde{\boldsymbol{q}}) \dot{\tilde{\boldsymbol{q}}}(t) \leq \xi \frac{d-d_s}{d_i-d_s}-\hat{\boldsymbol{n}}_{o r_l}^{\top} \dot{\boldsymbol{p}}_{o_l}(t) \eqno{(13)}
$$
where $\tilde{\boldsymbol{q}}$ represents the subset of joint variables that affect the distance to the obstacles, denoted by $\boldsymbol{J}_{d}$.  
$d_i$ for the activation distance of the controller and 
$d_s$ for the safe stopping distance. The variable 
$l$ denotes the index of obstacles, and the presence of multiple obstacles leads to a substantial increase in the constraints derived from the equations, which significantly impacts the success rate of solving the optimization problem. In our SDF-SC framework, obstacles are not represented as individual geometric shapes but are instead denoted as a collection of point clouds. Consequently, we can improve the inequality constraints as follows:
$$
\frac{\partial D}{\partial \tilde{\boldsymbol{q}}}
\dot{\tilde{\boldsymbol{q}}} \leq \xi \frac{D-d_s}{d_i-d_s}- \dot{\boldsymbol{p}}-s  \eqno{(14)}
$$
where $\dot{\boldsymbol{p}}$  represents the maximum velocity of the detected obstacle points, and $s$  denotes the self-collision boundary, which is utilized to constrain the configurations to prevent self-collisions.

\begin{figure}[tbp]
\begin{minipage}[b]{.5\linewidth}
    \centering
    \subfigure[]{\label{fig:Reactive Control}\includegraphics[width=1\linewidth]{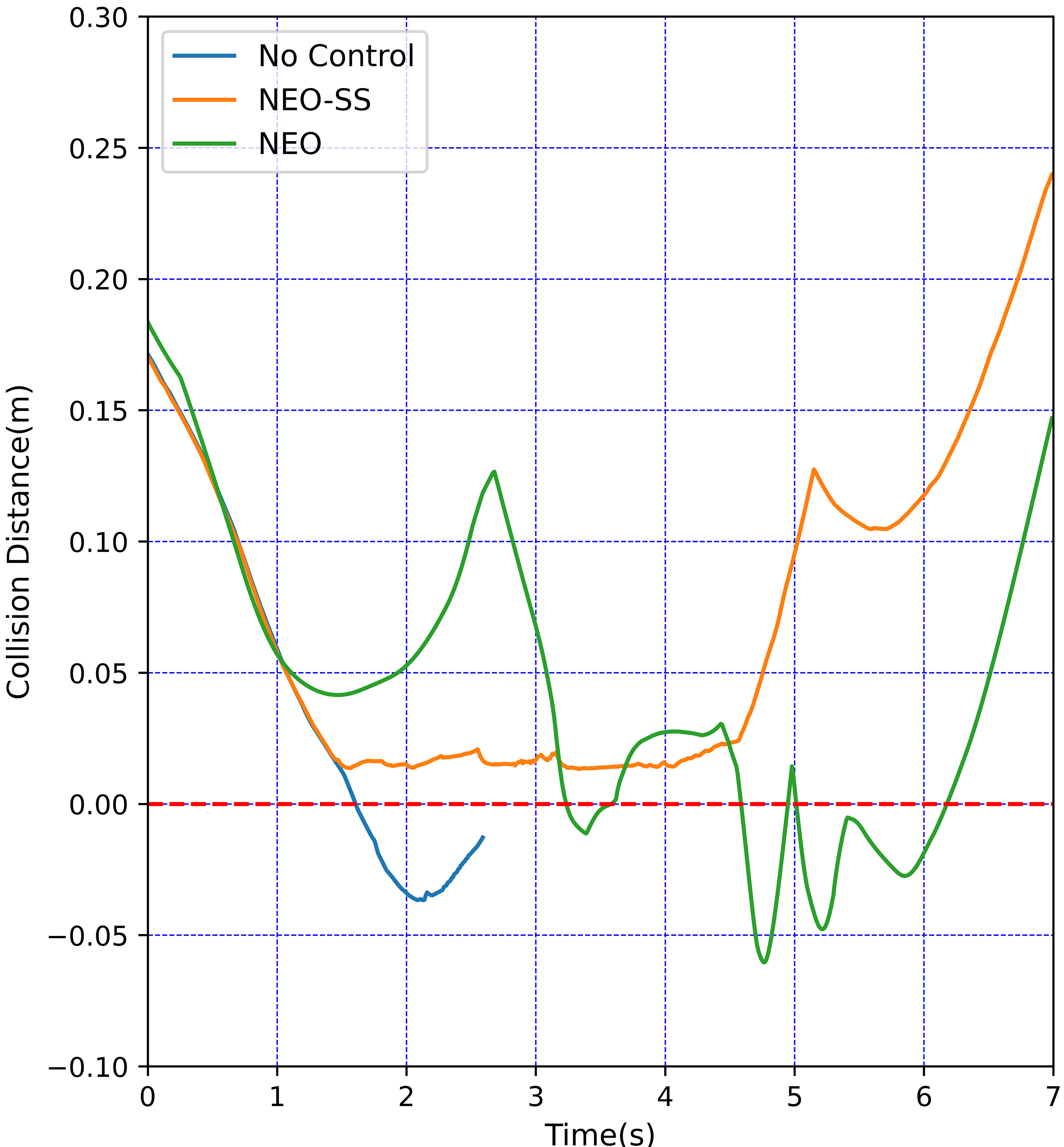}}
\end{minipage} 
\hfill 
\begin{minipage}[b]{.47\linewidth}
    \centering
    \subfigure[]{\label{fig:sim}\includegraphics[width=1\linewidth]{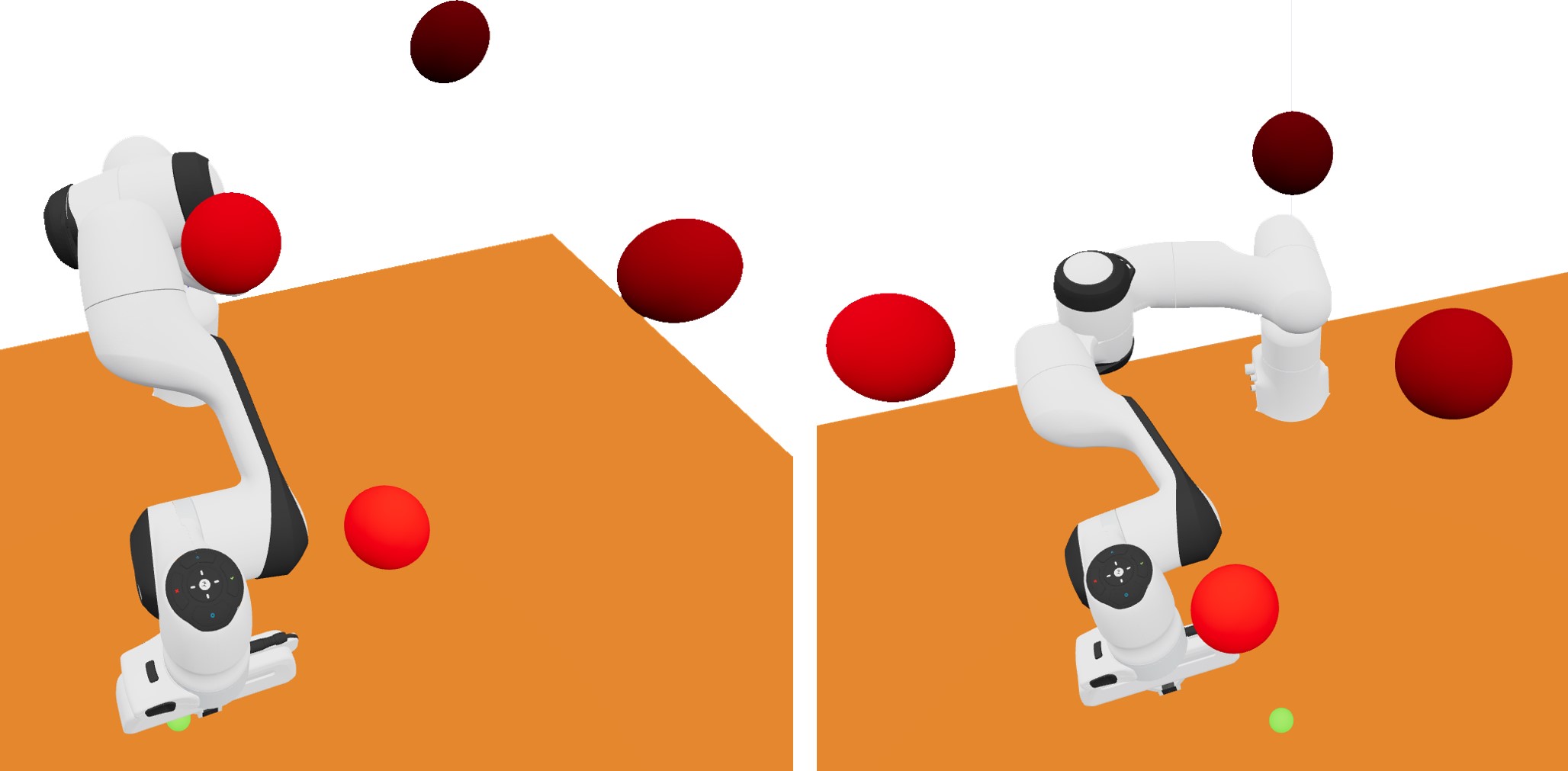}}
    
    \subfigure[]{\label{fig:real}\includegraphics[width=1\linewidth]{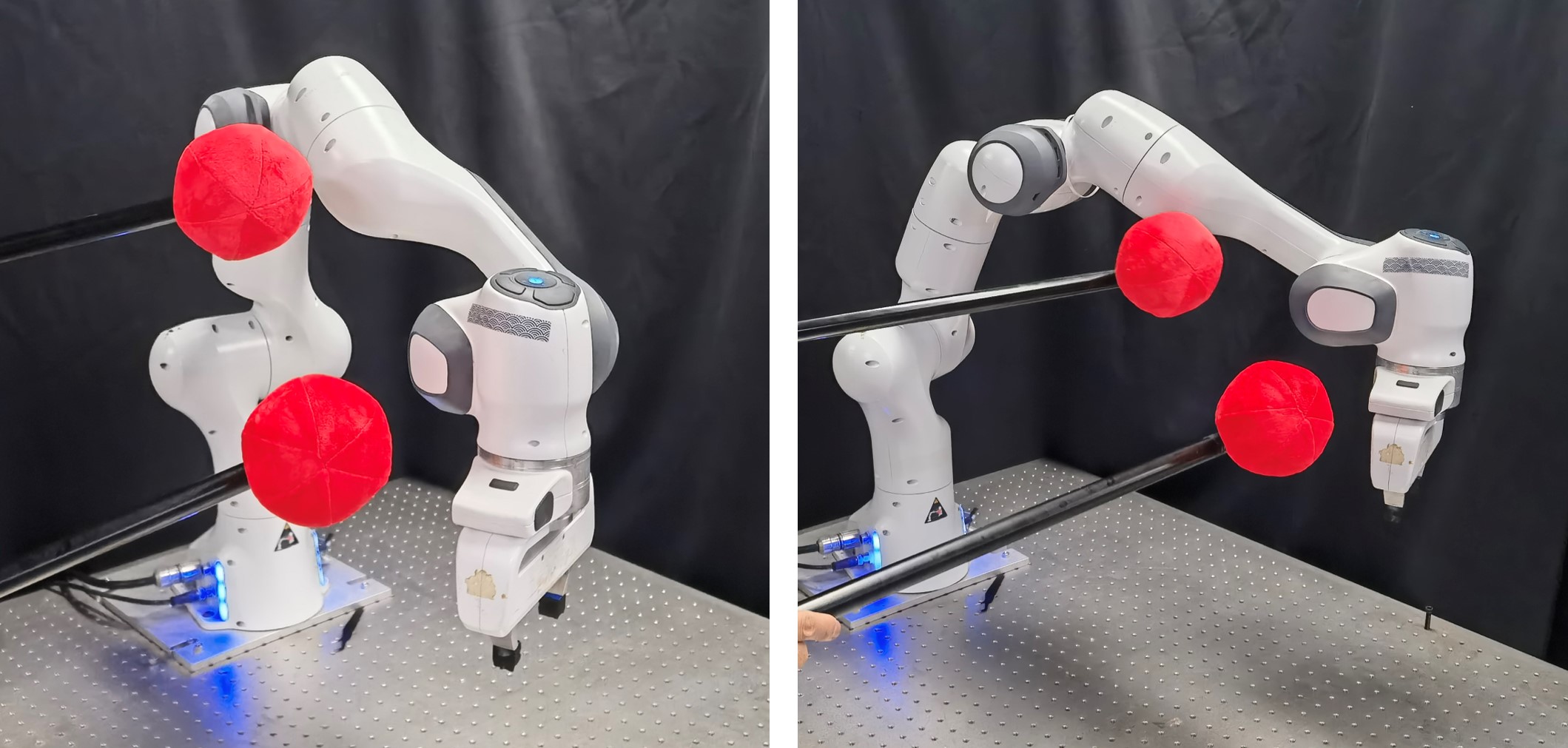}}
\end{minipage}
\vspace{-5pt}
\caption{(a) Collision distance during the movement period. (b) Simulation experiment. (c) Real robot experiment.}
\label{fig:NEO}
\vspace{-12pt}
\end{figure}

To assess the effectiveness of the improvements, we construct a simulation scenario as depicted in Fig. \ref{fig:sim}. In the scenario, multiple obstacle spheres are moving toward the robotic arm, which must complete the task of moving to a specified posture while avoiding these dynamic obstacles. In order to compare the effectiveness of the algorithm before and after the improvement of the inequality constraints, we set the adjustable parameters of NEO and NEO-SS to be consistent, with $d_i=$0.4m, $d_s=$0.02m, and $\xi=1.0$, and testing experiments are conducted under the same scenario. The experimental data is illustrated in Fig. \ref{fig:Reactive Control}. It is evident that NEO-SS exhibits reduced fluctuations, effectively keeping the distance well within the safety threshold. Conversely, NEO exhibits excessive fluctuations and results in a collision.

To evaluate the robustness of the reactive control algorithm, we randomize the positions and velocities of the obstacles within a certain range. The velocity of each obstacle fluctuates randomly around 0.2 m/s. We perform 50 experiments to ascertain the per-step computational latency of two control algorithms and to evaluate the efficacy of collision avoidance in each experiment. The compiled data regarding per-step latency and success rates are articulated in Table \ref{tab:NEO}. The experimental data indicates that in complex scenarios involving multiple dynamic obstacles, the NEO-SS algorithm demonstrates superior robustness compared to the NEO algorithm.

As shown in Fig. \ref{fig:real}, we implement our reactive control algorithm NEO-SS on a robot arm platform. The platform comprises a Franka Emika robotic arm and a Realsense D435i camera mounted at a fixed location for detecting obstacles. The robotic arm endeavors to execute reactive control measures to avoid the introduced dynamical obstacles concurrent with its mission to reach the designated target. 
During the experimental procedure, the robotic arm exhibits a distinct aversion to obstacles. As the obstacles approach within a certain threshold, the arm articulates its joints to move away from the obstacles, endeavoring to maintain the collision distance within a safe margin. Once the obstacles exit the reactive controller's activation range, the arm resumes its trajectory toward the target posture, calculated based on the inverse kinematics solution. For demonstration videos in both simulation and real-world environments, please refer to the attached video files.
\begin{table}
    \centering
    \caption{Comparison between NEO and NEO-SS}
    \begin{tabular}{c|c|c}
    \hline
    Controller & NEO & NEO-SS\textbf{(ours)}\\
    \hline
      Average time  & \textbf{3ms} & 6ms\\
    \hline
       Success rate  & 58\% & \textbf{86\%}\\
    \hline
    \end{tabular}
    \label{tab:NEO}
\end{table}

\section{CONCLUSIONS}

In this paper, we improved the framework of learning SDF based on the forward kinematics of articulated robots, utilizing multiple extremely lightweight networks in parallel to more efficiently approximate SDF. Additionally, we innovatively introduced self-collision detection into the framework, resulting in a collision detector that completely prevents collisions, termed SDF-SC. We then evaluated the detector’s performance in both self-collision and external collision detection, achieving highly satisfactory results in both domains. Furthermore, we utilized the SDF-SC as a collision constraint in trajectory optimization, enabling the robot to minimize path cost during planning while ensuring collision-free operation. Lastly, we integrated SDF-SC into the design of a reactive controller, successfully achieving dynamic obstacle avoidance in complex environments.

There are aspects of our approach that offer potential for further improvement. 
In the future, we plan to extend our framework to dual-arm robots to further enhance their capabilities.
\addtolength{\textheight}{-0cm}   





\end{document}